\documentclass{article} 
\usepackage[preprint]{colm2025_conference}

\usepackage{microtype}
\usepackage{hyperref}
\usepackage{url}
\usepackage{booktabs}
\usepackage{times}
\usepackage{latexsym}
\usepackage{mdframed}
\usepackage{multirow}
\usepackage{multicol}
\usepackage{listings}
\usepackage{xcolor}
\usepackage{amsmath}
\usepackage{booktabs}
\usepackage{hyperref}
\usepackage{titlesec}
\usepackage{amssymb}
\usepackage{tcolorbox}

\usepackage{lineno}

\definecolor{darkblue}{rgb}{0, 0, 0.5}
\hypersetup{colorlinks=true, citecolor=darkblue, linkcolor=darkblue, urlcolor=darkblue}

\title{Multilingual Question Answering in Low-Resource Settings: A Dzongkha-English Benchmark for Foundation Models}

\author{Md. Tanzib Hosain \& Rajan Das Gupta \\
Department of Computer Science\\
American International University-Bangladesh\\
Dhaka, Bangladesh \\
\texttt{\{20-42737-1,18-36304-1\}@student.aiub.edu} \\
\And
Md. Kishor Morol \\
Department of Computing and Information Science \\
Cornell University \\
New York, United States of America \\
\texttt{mmorol@cornell.edu} \\
}

\begin{document}

\ifcolmsubmission

\fi

\maketitle

\begin{abstract}
In this work, we provide DZEN, a dataset of parallel Dzongkha and English test questions for Bhutanese middle and high school students. The over 5K questions in our collection span a variety of scientific topics and include factual, application, and reasoning-based questions. We use our parallel dataset to test a number of Large Language Models (LLMs) and find a significant performance difference between the models in English and Dzongkha. We also look at different prompting strategies and discover that Chain-of-Thought (CoT) prompting works well for reasoning questions but less well for factual ones. We also find that adding English translations enhances the precision of Dzongkha question responses. Our results point to exciting avenues for further study to improve LLM performance in Dzongkha and, more generally, in low-resource languages. We release the dataset at: \href{https://github.com/kraritt/llm_dzongkha_evaluation}{https://github.com/kraritt/llm\_dzongkha\_evaluation}.
\end{abstract}

\section{Introduction}
GPT-4 and other LLMs have shown notable proficiency in managing intricate natural language tasks including question-answering and reasoning. Recently, a great deal of study interest has been generated by these developments \cite{Bang2023, Liu2023}. Medium- and low-resource languages have been neglected despite these advancements, since the majority of attention has been on English and other high-resource languages. This problem is made worse by the fact that the majority of benchmarks used to assess LLM performance are only available in English and a few other languages. Because assessment frameworks lack linguistic variety; it is difficult to compare and evaluate LLMs' performance in low-resource language environments.

As LLMs like ChatGPT are more included into educational institutions, the severity of this problem is increased \cite{KhanAcademy2023}. The need for linguistically inclusive AI development is highlighted by the possibility that these technologies might exacerbate global inequalities in access to digital resources and education if they do not perform fairly in non-English languages like Dzongkha.

This study fills this gap by concentrating on Dzongkha, a language that is severely underrepresented in Natural Language Processing (NLP) research while being spoken by approximately 600K people worldwide \cite{Zeidan2023}. The following contributions are ours:

\begin{itemize}
    \item A dataset, DZEN, covering over 5K scientific questions from Bhutan’s national curriculum for middle and high school exams, covering bilingual Dzongkha-English question-answering. It includes factual, application-based, and multi-step reasoning questions across various scientific subjects.
    \item DZEN enables direct LLM performance comparison across languages by aligning Dzongkha and English questions, ensuring a fair assessment and exposing structural weaknesses in low-resource language proficiency.
    \item Analysis of top LLMs reveals significant performance gaps between English and Dzongkha. To enhance Dzongkha performance, we explore context-specific prompting techniques, including CoT and hybrid translation-augmented prompts.
\end{itemize}

\section{Related Work}
\subsection{Benchmark for Multilingual Reasoning}
There are many English benchmarks that evaluate the reasoning skills of LLMs, such as CommonsenseQA \cite{Talmor2019}, HellaSwag \cite{Zellers2019}, CosmosQA \cite{Huang2019}, and COPA \cite{Roemmele2011}. X-COPA \cite{Ponti2020} is extensively used for non-English languages, and more recent initiatives such as \cite{Doddapaneni2023} have translated it into a number of Indic languages, while Dzongkha is still not included. Other benchmarks, such as BIG-Bench Hard \cite{Suzgun2022}, choose 23 difficult problems from BIG-Bench \cite{Srivastava2022} to assess reasoning in areas like multi-step arithmetic and logical deduction.

The majority of recent datasets that include academic test questions have been English-focused. Examples include the ARC dataset \cite{Clark2018} for scientific questions, the MMLU \cite{Hendrycks2020} for issues ranging from elementary school to college level, and MATH \cite{Hendrycks2021} for competitive mathematics and GSM8k \cite{Cobbe2021} for elementary school math problems. Work by \cite{Kung2023} and \cite{Choi2023} and studies like the GPT-4 technical report \cite{OpenAI2023} further benchmark LLMs on examinations like the AP and GRE. A famous example of a non-English exception is IndoMMLU \cite{Koto2023}. Resources for Dzongkha are very limited. Although it contains Dzongkha, the translated math dataset MGSM \cite{Shi2022}, which spans ten languages, lacks a more comprehensive academic perspective.

We provide DZEN, a collection of test questions taken from Bhutan's national school curriculum, in order to fill this need. DZEN is the first Dzongkha benchmark that covers a wide range of topics and question formats, allowing for a thorough assessment of LLMs in this underrepresented language.

\subsection{Multilingual Prompting}
A notable breakthrough in eliciting reasoning skills in large language models (LLMs) has been brought about by Chain-of-Thought (CoT) prompting \cite{Wei2022, Wei2023}. The usefulness of CoT was emphasized by \cite{Shi2022}, especially in improving mathematical thinking via step-by-step analysis in non-English languages. Their studies also showed that utilizing Google Translate or other similar tools to translate queries into English and then reasoning in English step-by-step often produces even better outcomes. \cite{Ahuja2023} used a similar approach. Cross-linguistic thought prompting, which entails translating the initial inquiry into English before doing CoT reasoning in English, was more recently developed by \cite{Huang2023}.

Nonetheless, a significant drawback of using English for CoT reasoning is that it lessens the usefulness of LLM outputs for users who do not know the language. This is especially important in educational contexts, as CoT often acts as a useful instrument for giving students perceptive guidance \cite{Han2023}. In this work, we show that, if an English translation of the questions is available, conducting CoT reasoning directly in the target language may enhance performance. For non-English speakers, this method guarantees increased accessibility and relevance, particularly in situations where language-specific reasoning is essential.

\section{DZEN Benchmark}
A similar set of multiple-choice English-Dzongkha scientific questions from exams in the eighth, tenth, and twelfth grades make up our dataset, DZEN. These questions are taken from Bhutan's national board examinations\footnote{\href{https://bcsea.bt/}{https://bcsea.bt/}} and are formally accessible in both Dzongkha and English, guaranteeing a high-quality parallel corpus. In educational research, this dual-language availability enables strong cross-linguistic analysis and application.

\begin{table}[htbp]
\scriptsize
    \centering
    \caption{Statistics of DZEN dataset grouped by various fields of subjects and levels.}
    \begin{tabular}{p{4cm}p{4cm}p{4.5cm}}
        \toprule
        \textbf{Group} & \textbf{\#Fields} & \textbf{\#Instances (\%)} \\
        \midrule
        \multicolumn{3}{c}{\textit{grouped by level}} \\
        Biology & 3 & 1004 (19.45\%) \\
        Chemistry & 3 & 1160 (22.48\%) \\
        Physics & 3 & 970 (18.79\%) \\
        Mathematics & 5 & 1794 (34.76\%) \\
        Science & 1 & 233 (4.51\%) \\
        \midrule
        \multicolumn{3}{c}{\textit{grouped by field}} \\
        12th Grade & 8 & 2857 (55.36\%)\\
        10th Grade & 5 & 1857 (35.98\%) \\
        8th Grade & 2 & 447 (8.66\%) \\
        \midrule
        \textbf{Total} & 15 & 5161 (100.00\%) \\
        \bottomrule
    \end{tabular}
    \label{tab:DZEN_statistics}
\end{table}

\subsection{Benchmark Design}
\subsubsection{Principles of Dataset Curation}
Apart from being accessible digitally, Bhutanese test questions are mostly printed. The ground truth for the questions was established by gathering physical test papers and using easily accessible answer guides. The questions and answers were digitized by three typists who were proficient in both Dzongkha and English. The information was converted into a digital format with minimum mistakes.

There are four possible answers for each multiple-choice question. The typists were told to use \LaTeX \vspace{1em} to format chemical formulae and mathematical equations in order to preserve correctness. The dataset does not include questions with figures. To further eliminate non-parallel questions, annotation mistakes, and low-quality items—such as those that were not parallel or had discrepancies between the English and Dzongkha ground truths—we further used certain heuristics. The next section goes into more depth about the specific criteria used to filter non-parallel questions.

\subsubsection{Principles of Parallel Corpus Creation}
Questions for Bhutanese national board examinations are usually written in English and then translated into Dzongkha. This method makes it easier to create a parallel dataset. The Dzongkha and English versions often have different question sequences, however; for example, the first question in the Dzongkha version can be the tenth question in the English version. To address this, we created a simple but very powerful algorithm for matching the Dzongkha and English queries.

First, the Google Translate\footnote{\href{https://translate.google.com/}{https://translate.google.com/}} API is used to translate each English question into Dzongkha. After that, we compute the cosine similarity between the Google translation and the official Dzongkha translation of the question for each topic. We do this by using the OpenAI $text-embedding-ada-002$ model. To make sure the Dzongkha question correctly matches its English equivalent, two native Dzongkha speakers who are also proficient in English personally check each one after the first matching. Furthermore, we eliminate inquiries in which the Dzongkha and English versions have different ground truths. Such disparities are usually caused by annotator mistakes or intrinsic problems with the questions themselves, according to manual examination.

Some English questions may have small grammatical mistakes or seem strange to native English speakers since these translations are often done by professors who have different degrees of English competence. We used GPT-4 to construct a grammar-corrected version of the English questions, with human supervision, in order to investigate the effect of these grammatical errors on model performance. As shown in Appendix \ref{app1}, grammar mistakes often have no discernible effect on model performance, according to our research.

\subsection{Precis of Dataset Properties}\label{dproperty}
A total of 5,161 questions in both English and Dzongkha are included in our suggested DZEN corpus. Some of the questions in this collection are from the 12th grade (55.36\% (2,857)), the 10th grade (35.98\% (1,857)), and the 8th grade (8.66\% (447)). The disciplines covered in the 12th grade section include biology, chemistry, physics, and mathematics. These are further divided into part I and part II according to subtopics. The dataset includes Biology, Chemistry, Physics, and Mathematics I and II for the tenth grade. Both science and mathematics are included in the dataset for the eighth grade, with science being a wide subject that covers all scientific subjects covered in the eighth grade. A comprehensive overview of the topic and the distribution of questions by grade is shown in Table \ref{tab:DZEN_statistics}.

\subsection{Precis of Dataset Categorization}
No matter the topic or grade level, the variety of question types in our collection necessitates a certain set of abilities. The questions are divided into three groups according to the abilities required to answer them as presented in Table \ref{tab1}.
\subsection{Style}

\begin{table}[htbp]
\centering
\scriptsize
\caption{Question categorization with respective description.}
\begin{tabular}{lp{10cm}}
\hline
\textbf{Category}              & \textbf{Description} \\ \hline
Factual Knowledge     & These inquiries just need the ability to remember fundamental facts, occasions, ideas, dates, or comparable details. There is no analysis or reasoning required to answer such queries. \\ \hline
Procedural \& Application & These problems fall into this category because they call for the use of a process, a well-known idea, or a formula in order to be solved. \\ \hline
Reasoning             & Several stages of analysis or reasoning are required to arrive at the right answer for questions in this category. \\ \hline
\end{tabular}
\label{tab1}
\end{table}

With the use of GPT-4 and the instruction given in Figure \ref{f20}, we categorize the questions into these categories. We note that the bulk of issues in chemistry and biology lean toward factual understanding. On the other hand, most mathematics issues call on procedural and application abilities, but sometimes thinking is also required. There is a fairly even mix of factual and procedural questions in Physics.

\section{Methodology}
\subsection{Experiment Setup}
We used the DZEN dataset to test a variety of open-source and proprietary LLMs to see how well they performed on our freshly created dataset. For both the Dzongkha and English datasets, we kept the system prompt in English in accordance with suggestions from earlier studies on ChatGPT \cite{Lai2023}. To reduce tokenization costs, especially for proprietary models, the majority of benchmark results reported in the main article were performed in a zero-shot setting \cite{Petrov2023}. Section \ref{app2} contains the specific results of further studies that were conducted in 3-shot and 5-shot situations.

\subsection{Experiment Models}  
The following models were used in order to evaluate the performance of existing LLMs on our dataset as shown in Table \ref{tab2}.
  
\begin{table}[htbp]
\centering
\scriptsize
\caption{Models used in tests.}
\begin{tabular}{lp{11cm}}
\hline
\textbf{Category}           & \textbf{Models} \\ \hline
Proprietary LLMs   & GPT-3.5 and GPT-4 \cite{OpenAI2023}, Claude 2.1 \\ \hline
Open-source LLMs   & LLaMA-2 7B and 13B \cite{Touvron2023}, Mistral 7B \cite{jiang2023mistral} \\ \hline
\end{tabular}
\label{tab2}
\end{table}

Notably, the majority of open-source models don't work well on Dzongkha. We only use the English version of the dataset for all open-source model trials because of this.

In this work, GPT-3.5 Turbo is used for the majority of the other experiments (such as ablations, prompting method investigation, and others). This decision was taken in order to minimize the considerable expenses that come with more sophisticated proprietary models, especially for Dzongkha \cite{Ahia2023, Petrov2023}. By choosing to benchmark on GPT-3.5, we guarantee a fair trade-off between price and value.

\subsection{Experiment Prompts}
We tested if certain prompting strategies could enhance LLMs' performance on our dataset. Recent research has shown the usefulness of the CoT prompt in improving reasoning tasks, thus we used it \cite{Wei2023}. We also ran trials without the CoT prompt for comparison. Appendix \ref{app3} contains comprehensive explanations of the prompts we utilized in our tests.  

The model was specifically told to use Dzongkha for CoT reasoning and refrain from using English in all Dzongkha experiments. This supports our main goal of making sure the model's output is helpful to users who speak Dzongkha, especially in educational settings. The steps might be generated in English first, and then translated into Dzongkha as an alternate method of creating CoT stages in Dzongkha. The "translationese" issue, in which translated phrases may seem awkward or grammatically wrong, may alter the original purpose or meaning, is why we decided against using this approach. 

\subsection{Experiment Evaluation Metrics}  
A manual examination of the model's outputs is part of the evaluation process to see if the final response is consistent with the ground truth. Note that the validity of intermediary reasoning\footnote{When comparing the intermediate CoT assessment with the final answer evaluation, we found that there are few cases in which the model correctly predicts the final answer when the CoT stages are incorrect.
} processes is not examined in our evaluation. Put another way, whether or not the intermediate steps in the reasoning process are correct, a solution is deemed correct if the outcome is accurate.  

\section{Experiment Results}  
\subsection{LLMs' Performance on DZEN}  
\subsubsection{All-subject Performance}  
The performance of different existing LLMs is assessed in this experiment for every subject in the DZEN dataset. For every model, the findings are shown in Table~\ref{tab:model_performance_v2}. It shows that proprietary LLMs perform much better than open-source models like LLaMA-2 and Mistral for English questions. When it comes to proprietary LLMs, GPT-4 performs the best, followed by GPT-3.5. Additionally, we see that LLM performance is generally stable across topics, with substantial declines in certain subjects, such 10th grade biology and 12th grade math II, indicating that these subjects may present more difficulties for LLMs. Similar patterns can be seen in Dzongkha questions, where GPT-4 continues to maintain a sizable score differential. Remarkably, Claude-2.1 performs far better in Dzongkha than it does in English, coming in considerably closer to GPT-3.5. The whole benchmark table is in Appendix \ref{app4}.

\subsubsection{Performance in English and Dzongkha}  
Two of the best-performing models, GPT-3.5 and GPT-4, are the subject of our investigation into the performance difference between English and Dzongkha questions. A significant performance difference between Dzongkha and English for GPT-3.5 is shown in Table~\ref{tab:model_performance_v2}. For GPT-4 (Table~\ref{tab:model_performance_v2}), on the other hand, the difference is much less, and performance in both Dzongkha and English increases dramatically in every topic. 

\subsection{Performance on CoT Prompting}\label{doescot}  
Prompting using CoT is often used to improve LLM performance on reasoning problems. We experimented across subjects and question categories to assess its efficacy on our dataset. 

\subsubsection{Subject-specific Performance}
\begin{figure}[htbp]
    \centering
    
    \includegraphics[width=.5\linewidth]{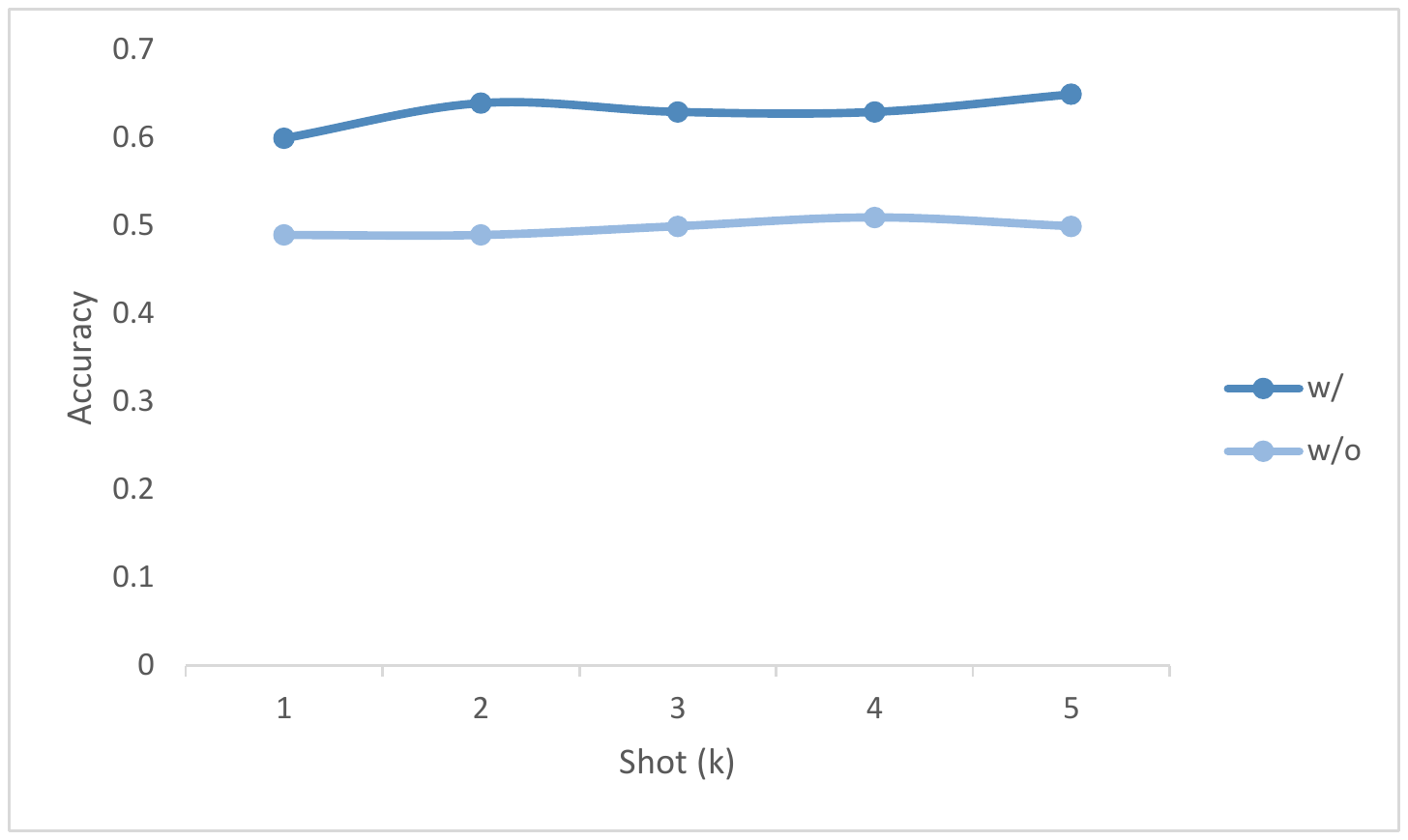}
    \caption{CoT reasoning average score in few-shot scenarios for the English GPT 3.5 Turbo. Note that w/ denotes with and w/o denotes without CoT.}
    \label{f5}
\end{figure}

\begin{figure}[htbp]
    \centering

    \includegraphics[width=.5\linewidth]{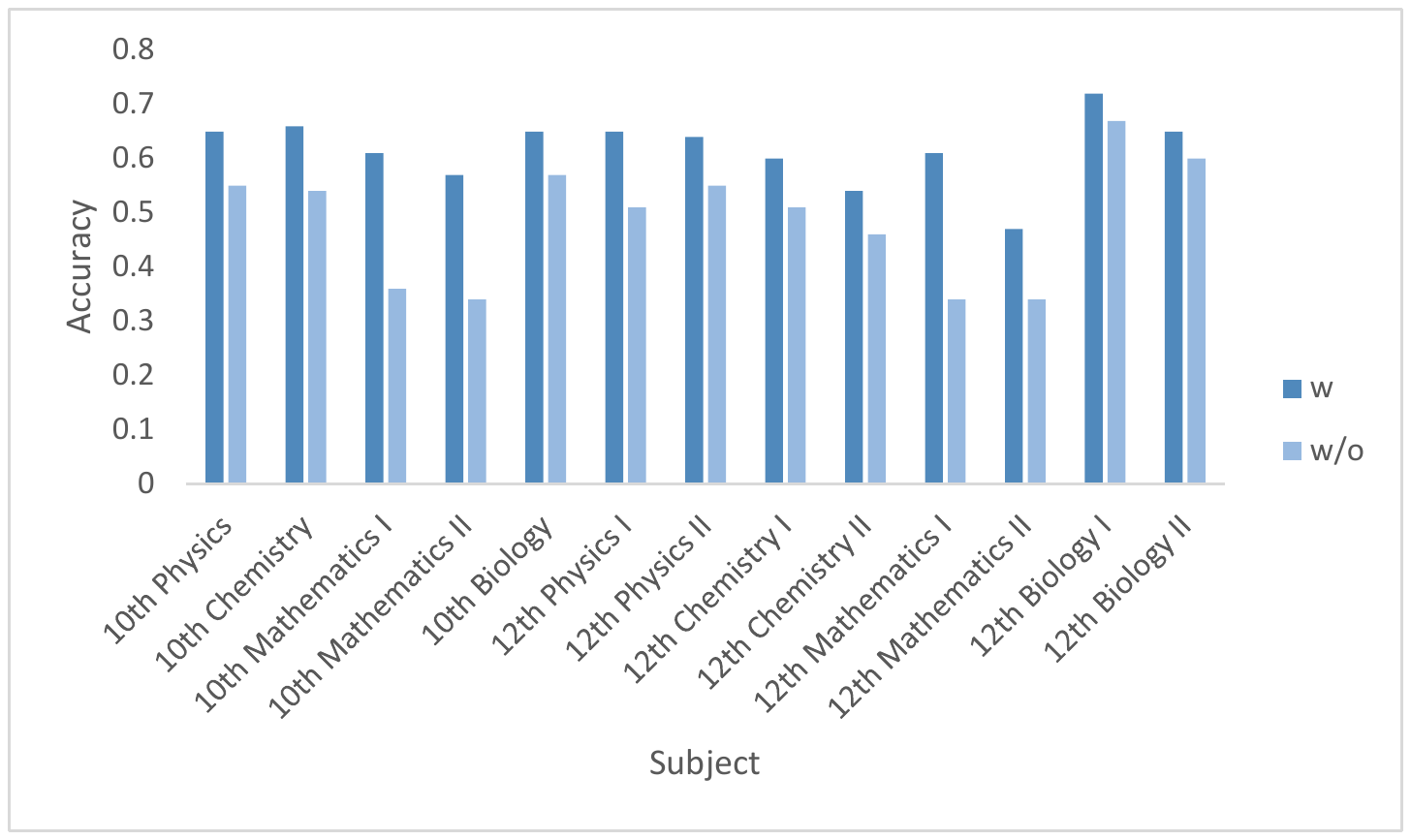}
    \caption{Subject-by-subject CoT performance in English. Note that w/ denotes with and w/o denotes without CoT.}
    \label{f6}
\end{figure}
Figure~\ref{f5} shows how CoT reasoning enhances GPT-3.5 performance for English questions throughout the dataset. A similar pattern is noted for Dzongkha; more information is given in Section~\ref{app2}. Figure~\ref{f6} illustrates that, when findings are split down by subject, CoT does not help all subjects equally. The distribution of question categories in each topic (Table \ref{tab:DZEN_statistics}) suggests that CoT helps math the most and biology the least. More specifically, there are more factual questions in biology and more reasoning and application-based questions in math.  

\subsubsection{Question-specific Performance}
\begin{figure}[htbp]
    \centering
    \includegraphics[width=.5\linewidth]{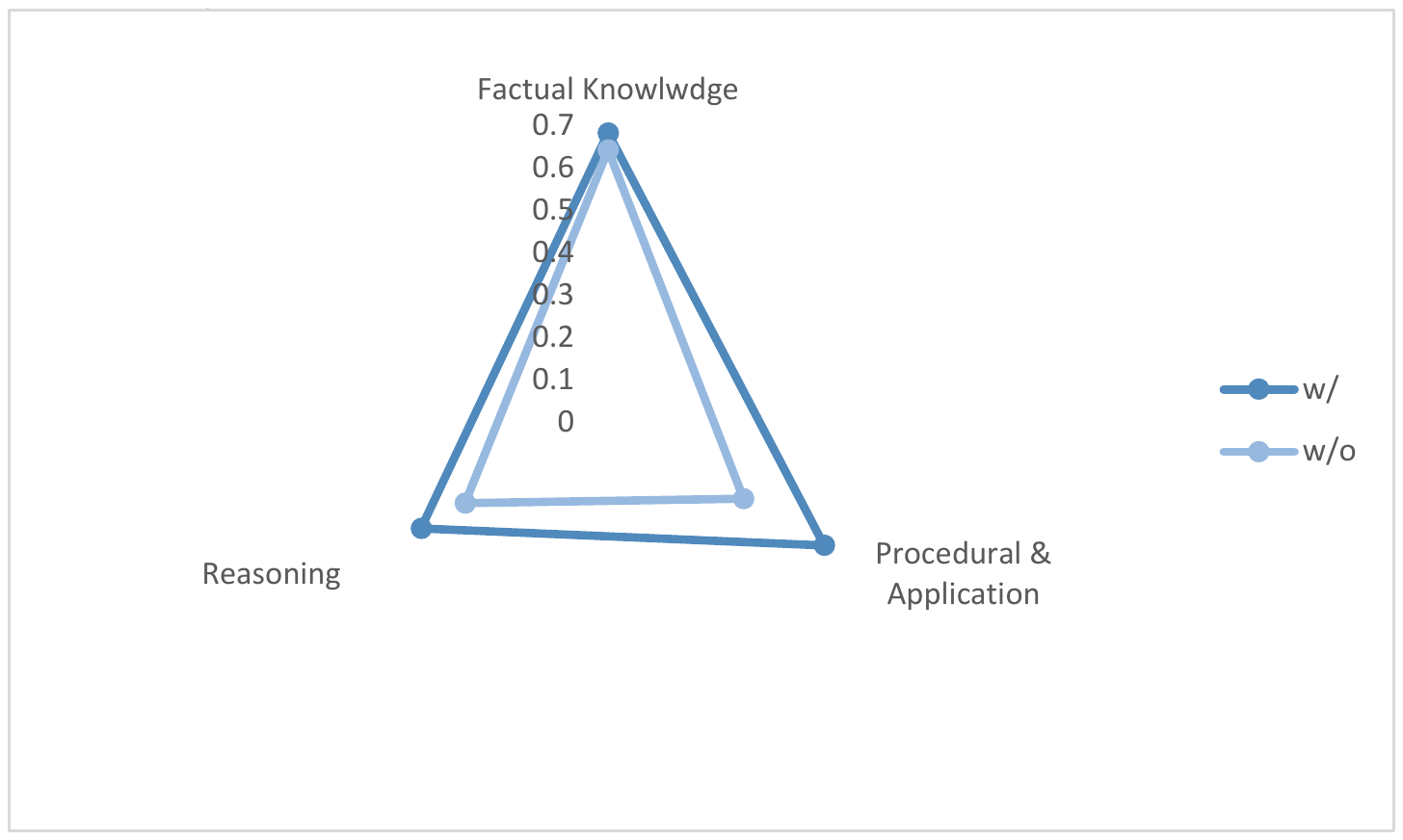}
    \caption{Performance summary in English by question type. Note that w/ denotes with and w/o denotes without CoT.}
    \label{f7}
\end{figure}
We examined GPT-3.5 performance by question category in order to further support our subject-based conclusions. CoT prompting had a strong positive impact on reasoning and application questions, as seen in Figure~\ref{f7}, with accuracy gains ranging from 12--22\%. The necessity for other strategies to increase performance in this area is highlighted by the factual questions, which show little progress. Notably, the improvements in application and reasoning questions via CoT for English questions still fall short of factual questions' performance. Section~\ref{app2.2} shows that Dzongkha has made the most progress in application questions, with only modest advances in factual and reasoning categories.

\section{Further DZEN Results}\label{app2}
\subsection{Few-shot Results}
\begin{figure}[htbp]
    \centering
    \includegraphics[width=.5\linewidth]{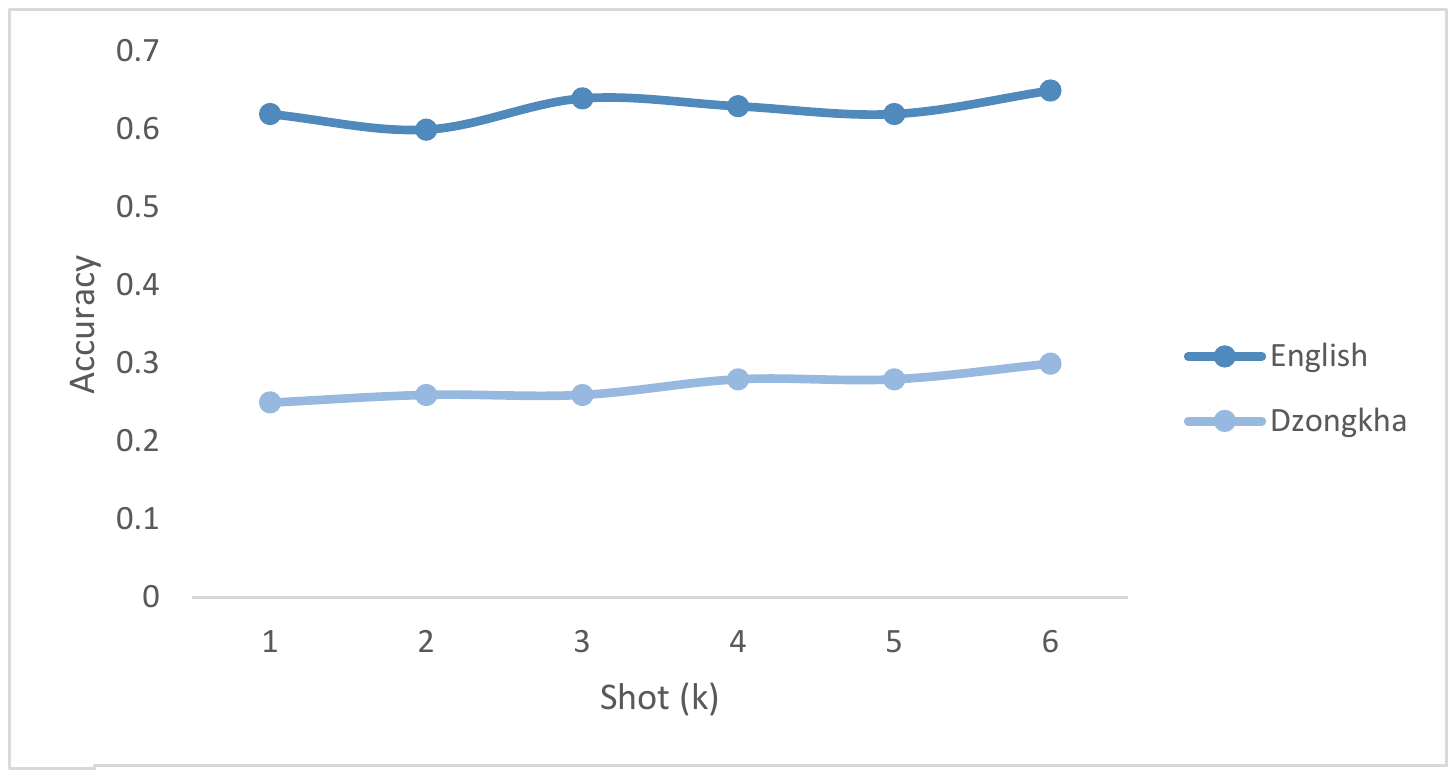}
    \caption{Dzongkha and English k-shot CoT prompted on GPT 3.5.}
    \label{f13}
\end{figure}
Although the preliminary trials showed that there is little difference between zero-shot and few-shot settings, we stick to zero-shot prompting throughout the paper (Figure \ref{f13}). therefore it is expensive to perform lengthy few-shot trials for proprietary models; this is made considerably more expensive in Dzongkha because to ineffective tokenization.

\subsubsection{Preparation of Few-shot Example}
Our categorization (factual, application, reasoning) as outlined in Section \ref{dproperty} was covered by at least the first three examples when creating the few-shot examples. The five examples also included every kind of question that typically appears in tests, including multiple-choice and multi-selection questions. Lastly, each topic's examples varied according to the subject.

\subsection{Impact of Prompting CoT}\label{app2.2}
\begin{figure}[htbp]
    \centering
    \includegraphics[width=.5\linewidth]{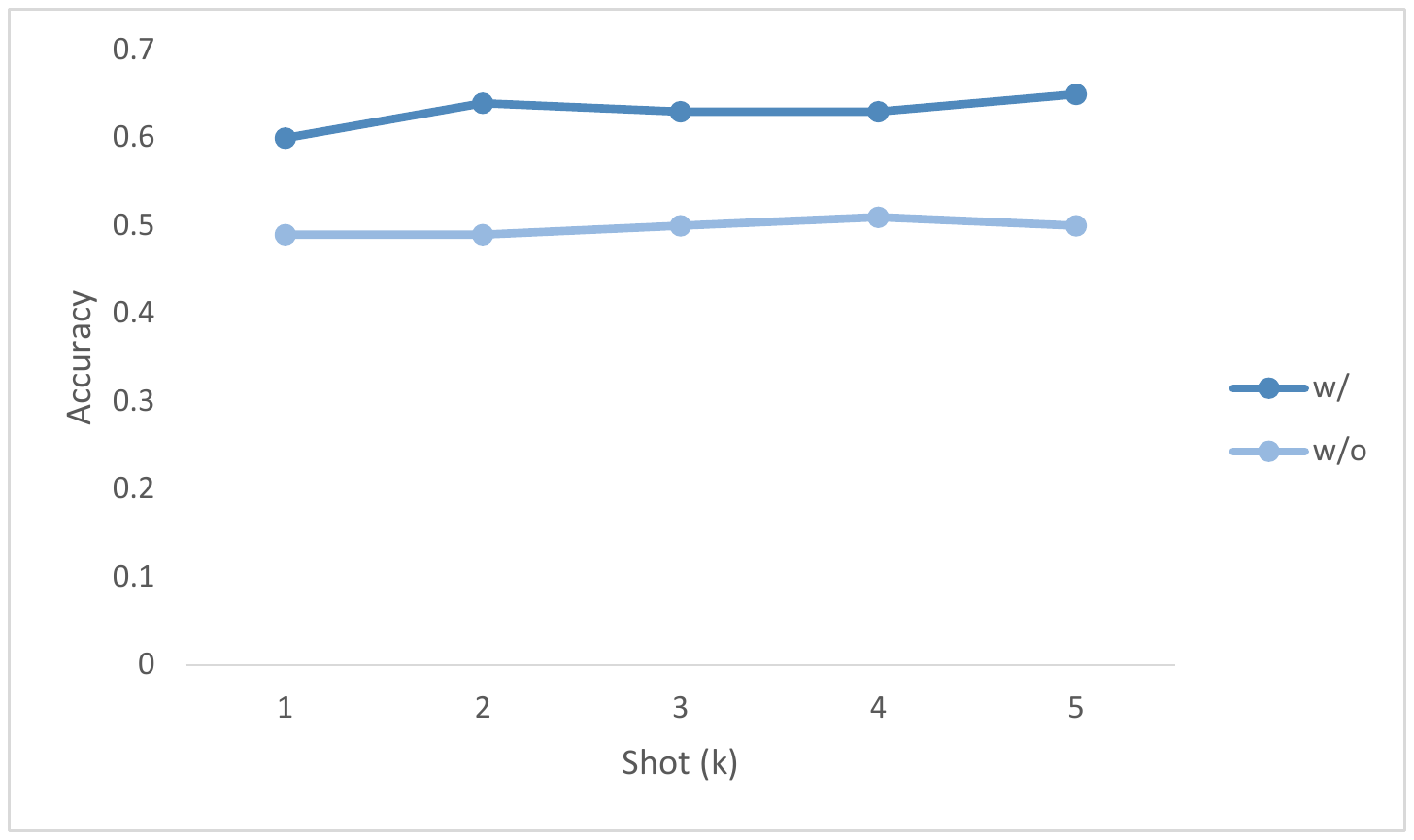}
    \caption{Impact of CoT reasoning on the GPT-3.5 for DZEN English throughout k-shot. Note that w/ denotes with and w/o denotes without CoT.}
    \label{f14}
\end{figure}

\begin{figure}[htbp]
    \centering
    \includegraphics[width=.5\linewidth]{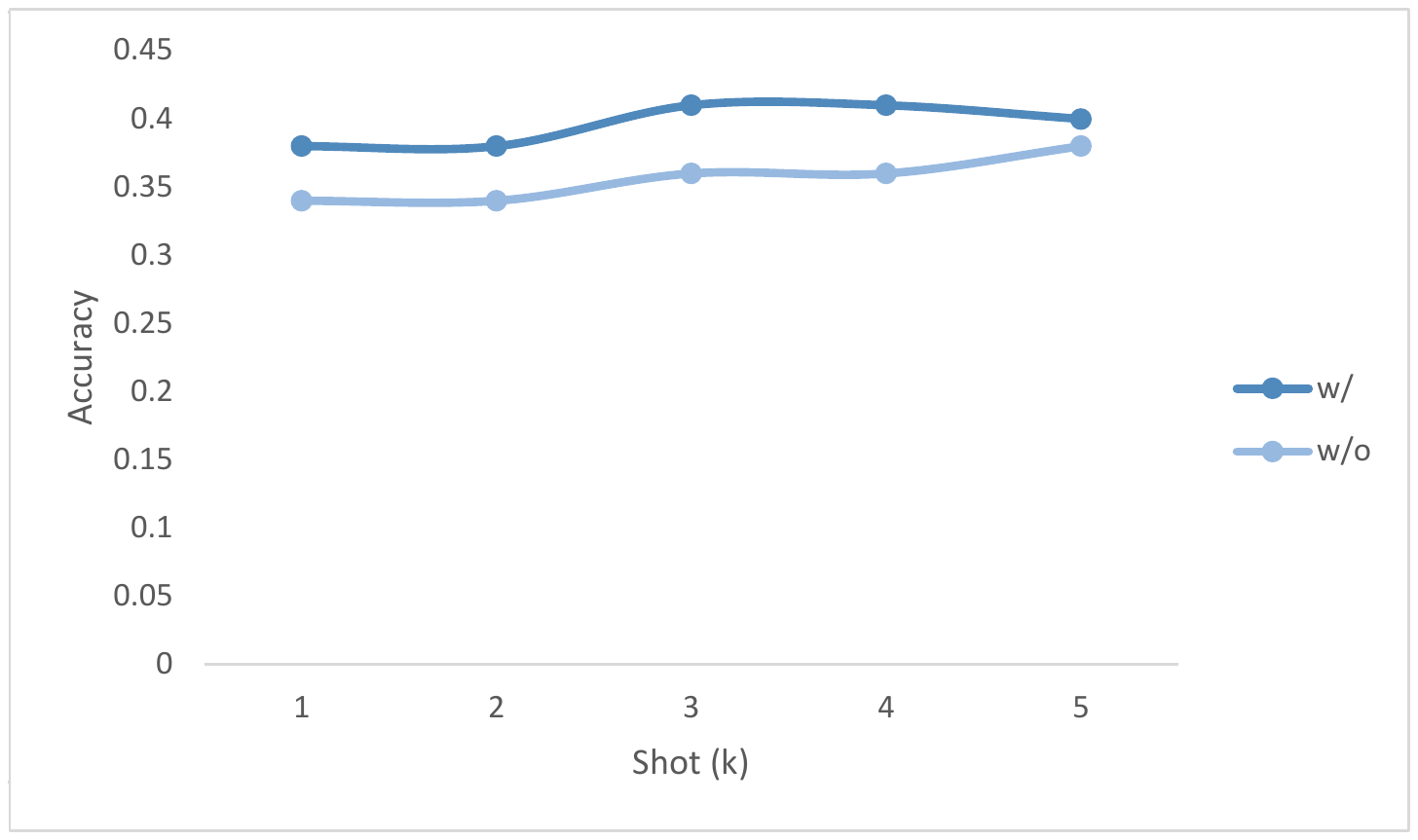}
    \caption{Impact of CoT reasoning on GPT-3.5 across k-shot. Note that w/ denotes with and w/o denotes without CoT.}
    \label{f15}
\end{figure}
For English, Figure \ref{f14} shows the CoT performance in 1 to 5-shot scenario. Using CoT always results in better performance. Similarly, GPT-3.5 in Dzongkha can be observed (Figure \ref{f15}).

\subsection{Subject-wise Breakdown of CoT}\label{app2.3}
\begin{figure}[htbp]
    \centering
    \includegraphics[width=.5\linewidth]{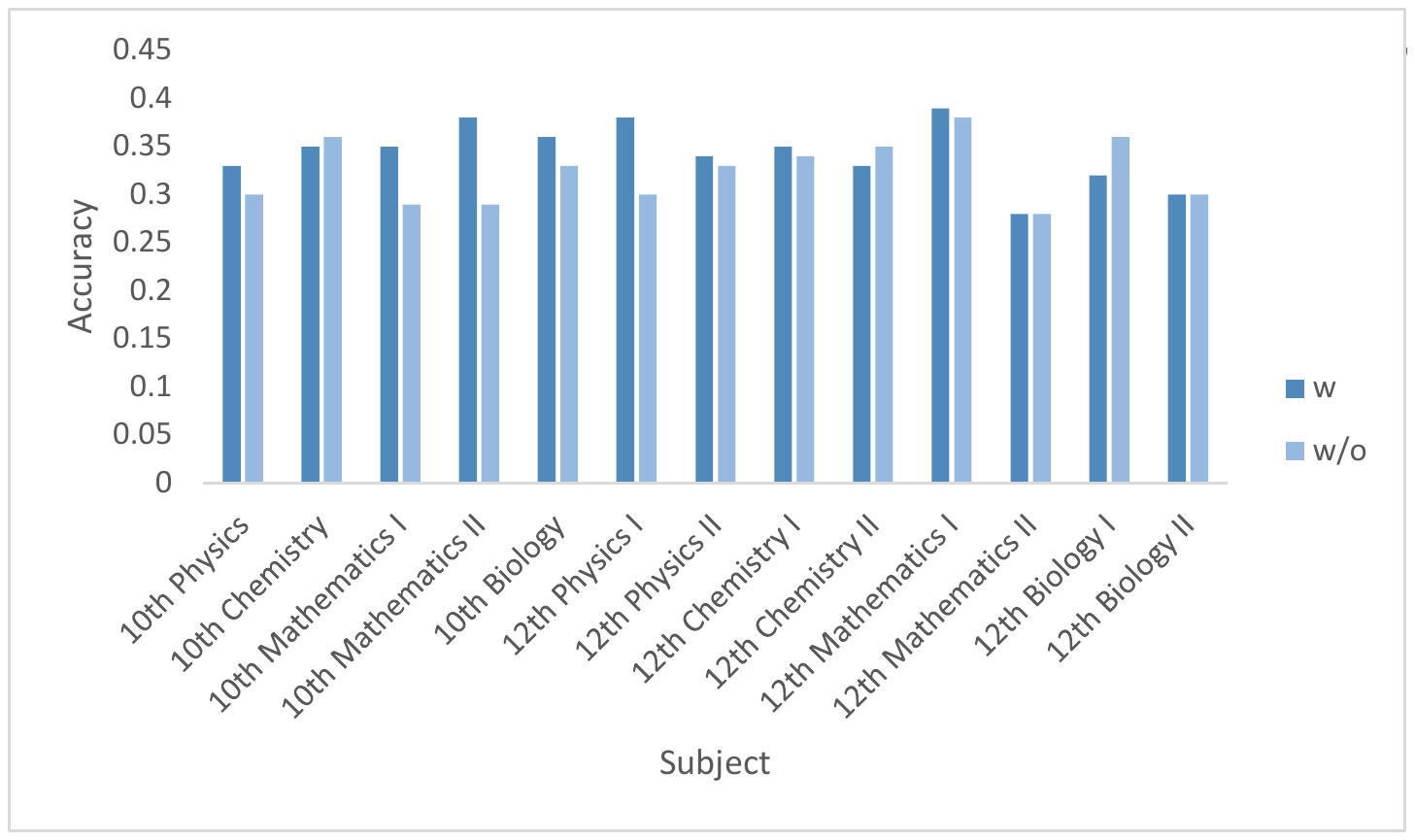}
    \caption{CoT performance summary in Dzongkha by subject. Note that w/ denotes with and w/o denotes without CoT.}
    \label{f16}
\end{figure}

\begin{figure}[htbp]
    \centering
    \includegraphics[width=.5\linewidth]{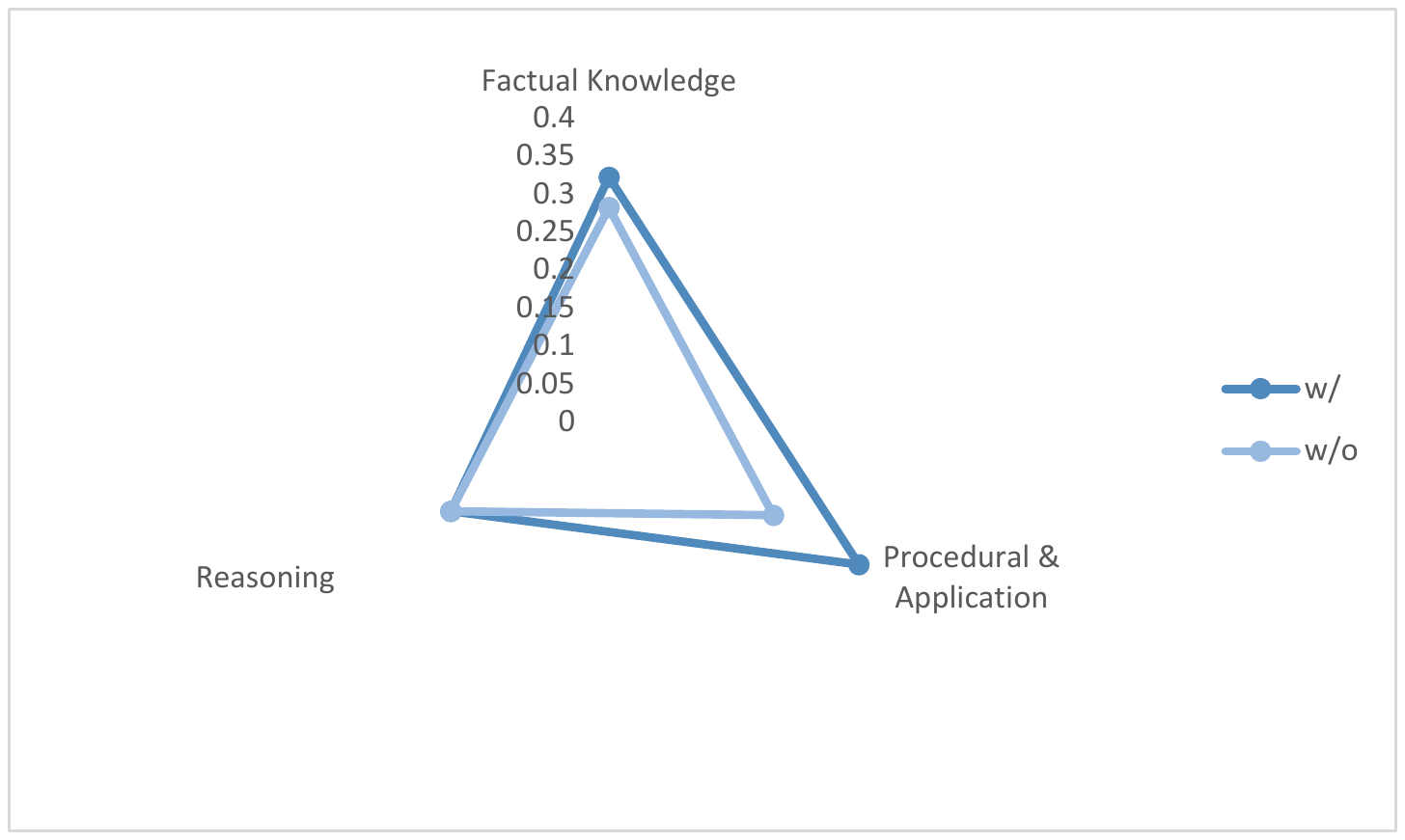}
    \caption{CoT performance summary in Dzongkha by question type. Note that w/ denotes with and w/o denotes without CoT.}
    \label{f17}
\end{figure}

We also note that completing CoT enhances Dzongkha performance, which is similar to what we saw in Section \ref{doescot}. The subject-wise breakdown of CoT performance for GPT-3.5 is displayed in Figure \ref{f16}.

Additionally, Figure \ref{f17} shows the CoT performance split for Dzongkha by question type.
Dzongkha performs similarly to English when it comes to reasoning problems. After analyzing the data, we hypothesize that this disparity results from the reasoning problems in some subjects—like physics— having smaller data sets. In addition, the model by default considers these questions more challenging to answer in Dzongkha than in English, as seen by the significantly lower base accuracy in Dzongkha.

\subsection{Performance of Translation Produced by LLM}
\begin{figure}[htbp]
    \centering
    \includegraphics[width=.5\linewidth]{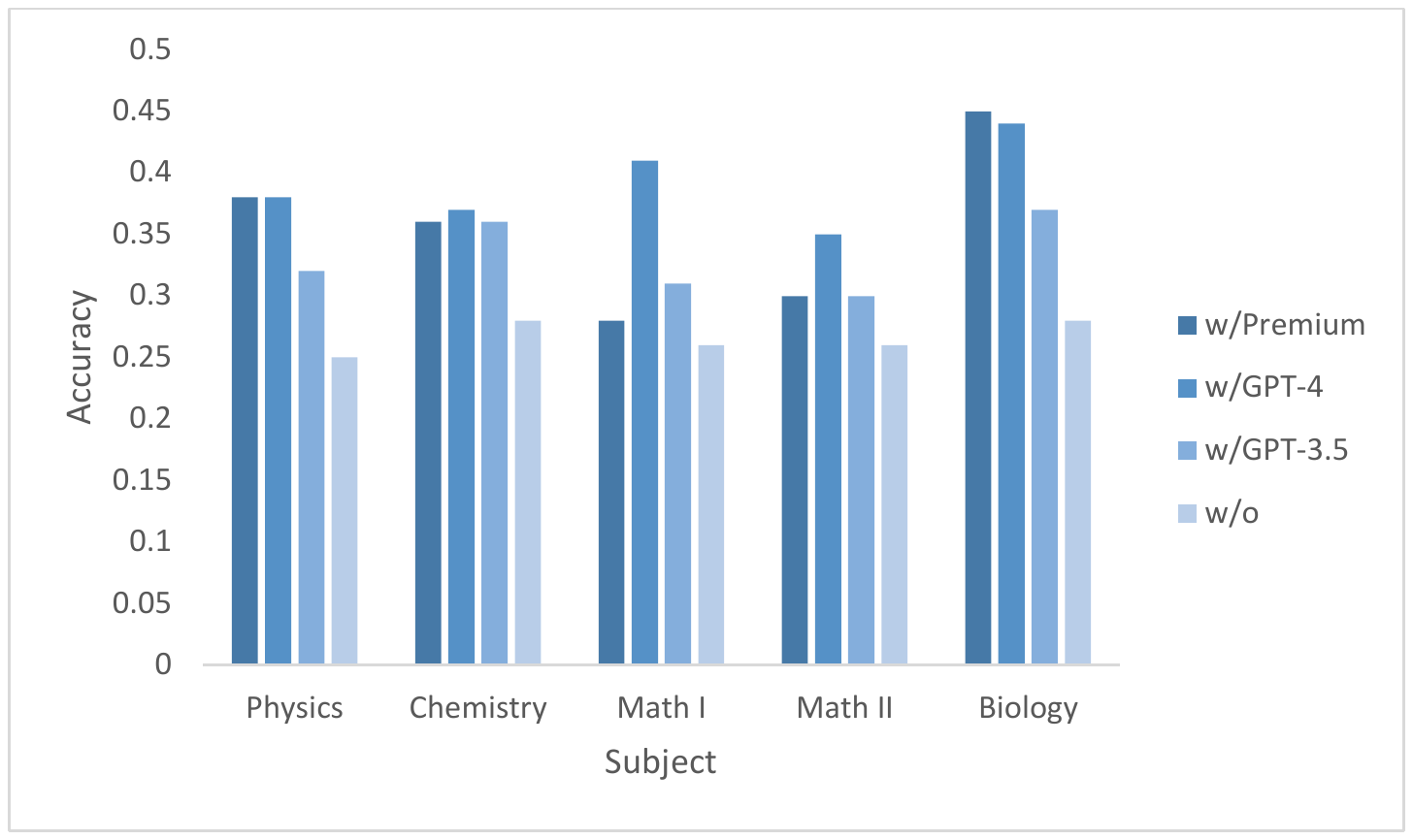}
    \caption{Consequences of adding an English version was produced by LLM in the adding experiment. w/o denotes the situation in which there was no English translation supplied.}
    \label{f18}
\end{figure}
Our translation adding prompt technique undoubtedly raises the question of whether machine translation can serve as a substitute when human-generated translations are unavailable.

Figure \ref{f18} demonstrates that adding an English translation produced by LLM performs equally well, and for some subjects even better than the original translation. Because Google Translate tends to break equations written in format, we solely employed GPT-3.5 and GPT-4 translations in this experiment and did not use Google Translate.

The result of the study raises the prospect of using these prompting strategies for a variety of multilingual tasks where it would be challenging to manually translate into English.

\section{Discussion}  
According to our benchmark results, there is still a lot of space for improvement in low-resource languages like Dzongkha for LLMs like ChatGPT. One noteworthy finding is that it is far more difficult to guarantee that the model outputs follow a specified structure for automated assessment in Dzongkha. Improving multilingual models' ability to follow instructions should be the main focus of future research.  

The significant performance difference between proprietary and open-source language models is another important conclusion. To compete with proprietary models, open-source models—which are more widely available in poorer nations—need to make substantial progress. This is necessary to make sure that no one group of people is excluded from the advantages of AI, especially LLMs. A recently published open multilingual LLM, Aya \cite{Ustun2024}, represents a promising step in this direction.  

We also showed that it is possible to use query translation to have LLMs answer in the target language while utilizing the benefits of high-resource languages, like English, in the inference pipeline. This method's advantage is that it does not need access to personally produced premium translations, as this study did; translations produced by the same model or more potent/specialized models can serve just as well. But the translation technique—whether it's a domain-specific, fine-tuned translation model or another skilled LLM—can affect performance. This result emphasizes how much more study is required to maximize and improve the usage of LLMs, especially for low-resource languages.

\section{Conclusion and Future Work}  
We presented DZEN, a locally derived dataset from Bhutan that includes exam questions at the middle and high school levels in both Dzongkha and English, in this study. Due to the parallel nature of our dataset, performance differences between the two languages may be compared more fairly. Even the best-performing models perform worse in Dzongkha than in English, according on benchmarking multiple LLMs on this dataset. Additionally, open-source models now fall below proprietary models by a wide margin.  

Additionally, we investigated if adding English translations to prompts could enhance Dzongkha question performance. Using the GPT-3.5 model, this strategy improved performance for the majority of participants in the DZEN dataset. Interestingly, additional dataset like Big-Bench-Hard also benefit from this enhancement. Future research aiming at enhancing LLM performance in low-resource languages, especially Dzongkha, is made possible by these discoveries.  

Results provide numerous encouraging avenues for further investigation. We developed a straightforward yet efficient prompting technique that greatly enhances performance on Dzongkha data since the results point to a possible gap in the current models' comprehension of Dzongkha language. We also used the Dzongkha Big-Bench-Hard dataset to show how effective this prompting method is. Future studies should examine how well these high-resource language prompting techniques transfer to different datasets and languages.

\bibliography{colm2025_conference}
\bibliographystyle{colm2025_conference}

\appendix

\section{Limitations}  
It is important to note that our work has various limitations. First, we removed figure-based questions from our dataset during curation, thus it mostly consists of text-based questions. Given that visual issues frequently call for more sophisticated thinking, this constraint could limit the breadth of our findings. Furthermore, because the questions are multiple-choice, there's a chance that models may skip some of the answers, particularly for factual questions that don't call for sophisticated thinking. For evaluating LLMs in Dzongkha, where resources for knowledge-intensive and question-answering duties are currently scarce, our dataset is a valuable starting point despite these drawbacks.  


\section{Further Tests}  
Here, we carry out further tests to see if using more effective prompting techniques will enhance Dzongkha performance. For Dzongkha, GPT-3.5 provides a decent mix between cost-effectiveness and capacity, and it was used for all of the trials discussed here.

\subsection{Performance on Adding English Translation} 
We speculate that two primary reasons for the poor performance of models in Dzongkha might be their lack of knowledge of Dzongkha scientific terms and possible challenges in deciphering non-Latin characters \cite{Lai2023}. We anticipate that giving the query an English translation will aid the model in comprehending the context. We experimented on a randomly chosen subset of 105 data points from each participant in the 10th-grade exam questions in order to test this hypothesis. Because they fall somewhere in the center of difficulty when compared to problems from the eighth and twelfth grades, the tenth grade questions were selected.  


\subsubsection{Findings}  
\begin{figure}[htbp]
    \centering
    \includegraphics[width=.5\linewidth]{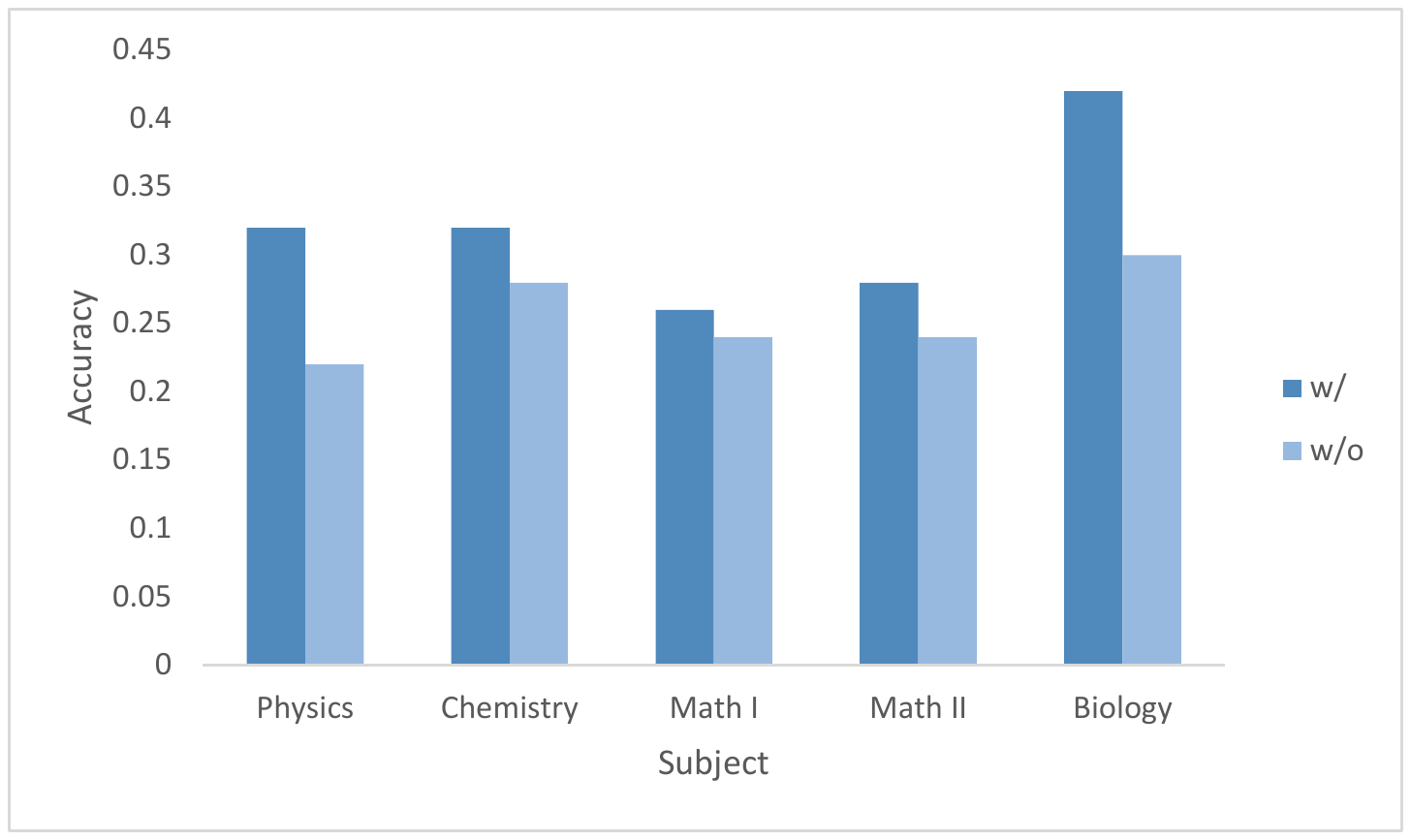}
    \caption{Answering questions in Dzongkha is made easier by including an English translation. Note that w/ denotes with and w/o denotes without including the translation. The model was requested to do CoT in Dzongkha.}
    \label{f9}
\end{figure}
Figure~\ref{f9} illustrates how adding English translations improves performance in every topic. The system prompts were always in English throughout our work. This experiment verified that the performance improvement is due to the attached translations by providing the questions in Dzongkha together with their English translations. In biology, where scientific vocabulary is widely used, the most improvement was shown. The gain in math, however, was not as noticeable. Additionally, we tested LLM-generated translations in place of human translations, and preliminary findings indicate that they could function similarly. Additional information can be found in Section~\ref{app2.3}.  

\subsection{Performance of Translation-appended Prompting Strategy to Different Datasets}  
To assess the suitability of our prompting approach on a variety of datasets, we expanded our trials to include Big-Bench-Hard. 

\subsubsection{Big-Bench-Hard Dzongkha}  
We used GPT-4 to create a Dzongkha version of the Big-Bench Hard (BBH-DZ) dataset for this experiment. We chose activities from Big-Bench Hard that demand reasoning and are relevant to Dzongkha since only 11\% of the DZEN tests require reasoning abilities. Due to the possibility of irregularities in the alphabetical order after translation, tasks such as word sorting were not included. Prompts for each challenge were iteratively created by two native Dzongkha speakers. 

When premium English translations were added, our tests revealed an average performance gain of 6.52\% in BBH-DZ, while GPT-4-generated translations produced an average improvement of 6.05\%. Additional information on the findings and task choices can be found in Appendix~\ref{app5}.

\section{Grammar Errors' Impact on DZEN}\label{app1}
\begin{figure}[htbp]
    \centering
    \includegraphics[width=.5\linewidth]{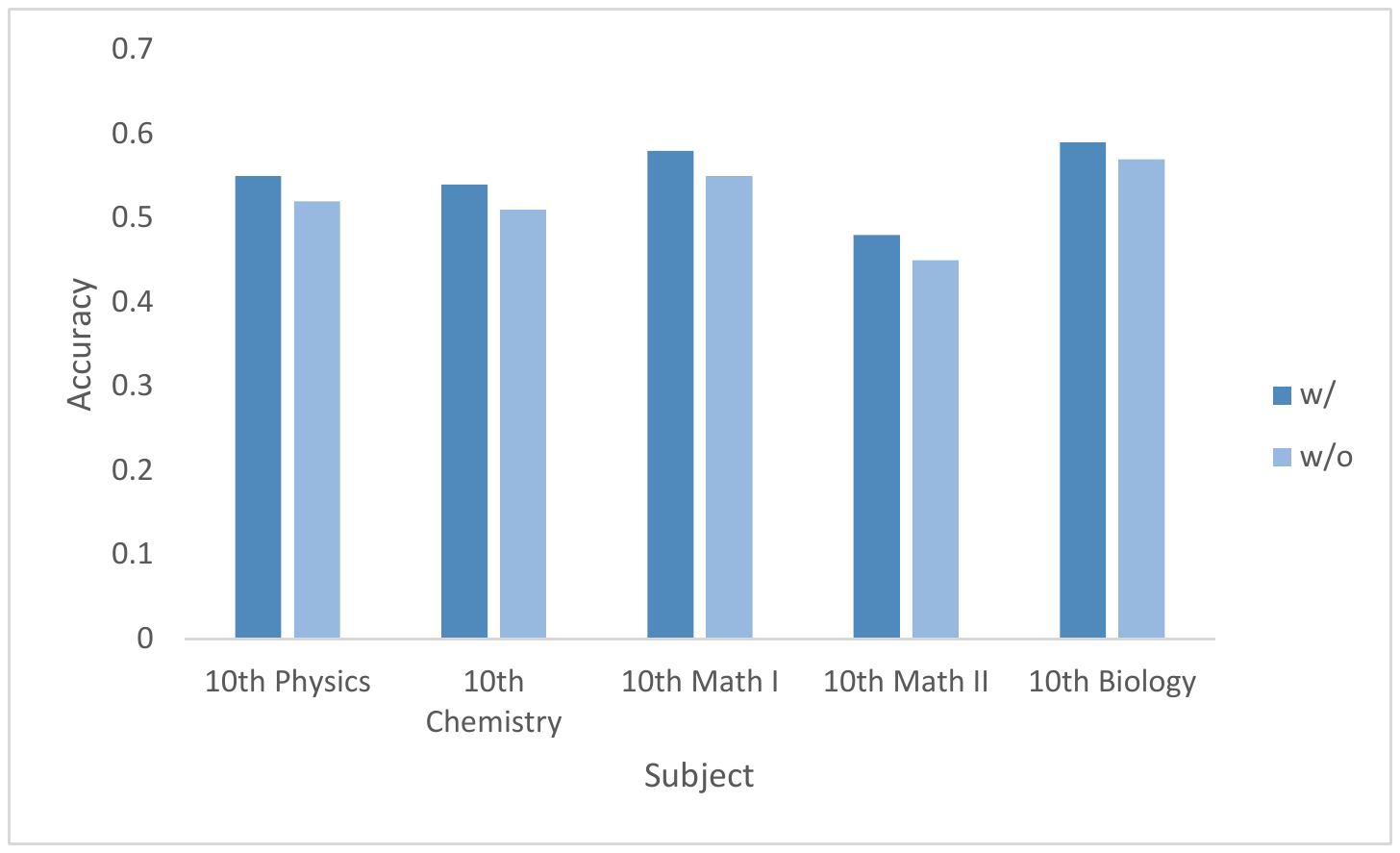}
    \caption{Grammatical errors' effects on GPT 3.5 performance. Note that w/ denotes with and w/o denotes without including grammar fixed.}
    \label{f12}
\end{figure}
In order to observe the impact of minor grammatical errors and strange English translations, we chose 115 questions from every 10th grade subject. We asked GPT-4 to correct any grammatical errors and strange translations. After reviewing the findings, a native Dzongkha speaker who was fluent in English made the required revisions\footnote{We did not utilize this procedure to repair grammar for the whole dataset since GPT-4 frequently changes the question's original meaning while fixing grammar.} to the GPT-4.
Figure \ref{f12} shows how well GPT-3.5 performed on this portion of the dataset. There is very little difference between the version with and without grammatical faults, and a manual examination of Figure \ref{f12} showed a little disparity that resulted from the stochastic nature of GPT-3.5 rather than any grammar issues.

\section{Results for Intermediate CoT}
\begin{table}[htbp]
\scriptsize
\caption{Test of correctness of CoT procedures and the final answer. Note that w/ denotes with and w/o denotes without including English translation. The intermediate levels of reasoning were completed in Dzongkha in each instance.}
    \centering
    \begin{tabular}{p{4.5cm}p{4cm}p{4cm}}
        \toprule
        & \multicolumn{2}{c}{Final Answer} \\
        \cmidrule(lr){2-3}
        & $\checkmark$ & $\times$ \\
        \midrule
        \textbf{w/} & & \\
        \hline
        CoT $\checkmark$ & 25 & 3 \\
        CoT $\times$ & 2 & 28 \\
        \midrule
        \textbf{w/o} & & \\
        \hline
        CoT $\checkmark$ & 18 & 0 \\
        CoT $\times$ & 3 & 32 \\
        \bottomrule
    \end{tabular}
    \label{tab3}
\end{table}
Table \ref{tab3} displays the results of our human evaluations of a subset of the responses. The main findings are that the model's final response is incorrect when the CoT is incorrect and correct when the CoT is correct. There are very few cases where the model inadvertently uses the incorrect CoT steps to arrive at the correct answer.

\section{Further Tests on BIG-Bench-Hard}\label{app5}
\subsection*{Alteration of Prompt}
We explore four possible variations of the prompting method in Table \ref{tab4}.
\begin{table}[htbp]
\scriptsize
    \centering
    \caption{Prompt variations.}
    \begin{tabular}{p{5cm}p{8.5cm}}
        \toprule
        \textbf{Condition} & \textbf{Description} \\
        \midrule
        Dzongkha & Asking only in Dzongkha. \\
        Dzongkha + English (GPT-3.5) & Asking in Dzongkha and appending the English translation done by GPT-3.5. \\
        Dzongkha + English (GPT-4) & Asking in Dzongkha and appending the English translation done by GPT-4. \\
        Dzongkha + English (Premium) & Asking in Dzongkha and appending the English version from the original data. \\
        \bottomrule
    \end{tabular}
    \label{tab4}
\end{table}

\subsection{Task Selection}
From the 23 datasets officially published by BIG-Bench Hard, we selected 14 tasks that can have equivalence in Dzongkha\footnote{There isn't a direct "translation" in Dzongkha for certain BBH tasks, such as manipulating English alphabet letters.}. The tasks we selected for our work included: Reasoning About Colored Objects, Web of Lies, Multistep Arithmetic, Navigate, Object Counting, Penguin in a Table, Causal Judgement, Date Understanding, Disambiguation QA, Formal Fallacies, Logical Deductions Five, Seven, Three, and Multistep Arithmetic.

GPT-4 was used to translate these 14 tasks into Dzongkha, with three human-annotated examples serving as prompts for each task. Two Dzongkha speakers iteratively created the prompts to reflect the specifics of each job.

\subsection{Findings}
The outcomes of adding English translations to Big-Bench-Hard experiments are displayed in Table \ref{r101}. The performance was marginally harmed in three instances, while it was beneficial in seven of the fourteen. Performance was essentially unchanged in the other four circumstances.

\section{Experimental Prompts}\label{app3}
This section contains all the original prompts used for all the experiments.
\subsection{Prepared DZEN Questions with Correct Grammar}
We use the prompt in Figure \ref{f21} to correct the grammar of the original English questions using GPT-4.

\subsection{Dataset Categorization for DZEN}
As previously stated in Section \ref{dproperty}, we used GPT-4 to categorize our dataset questions into three groups: Factual Knowledge, Procedural \& Application, and Reasoning. To classify them, we utilize the prompt provided in Figure \ref{f20}.

We use a zero-shot strategy with the categorization prompt for every question in the dataset. In tabular visualization, the subject-wise question category is represented by Table \ref{tab:DZEN_question_count}.

\subsection{Prompting CoT for DZEN Zero-shot Benchmark}
The prompt for zero-shot benchmarking with the proprietary models is displayed in Figure \ref{f22}.

\subsection*{Prompting CoT for DZEN Few-shot Benchmark}
We use the prompt in Figure \ref{f23} to do few-shot benchmarking using the proprietary models.

\subsection{Translation Appended Benchmark Prompt}

In the translation-append experiment, as explained in Appendix \ref{app5}, we limit the model to only reasoning in English by using the question displayed in Figure \ref{f24}.

\section{Statistics that Benchmark}\label{app4}
Benchmark results for the types and datasets we used for our experiments are shown in this section.
\subsection{DZEN Zero-shot Benchmark}
The zero-shot benchmark results on DZEN are displayed in Table \ref{tab5}.

\subsection{Benchmark for DZEN Few-shot with and without CoT Reasoning}
The few-shot benchmark results on DZEN with and without CoT reasoning are displayed in Table \ref{tab6}.

\subsection{Benchmark for BIG-Bench-Hard Zero-shot}
The zero-shot benchmark results on a few chosen reasoning-based BIG-Bench-Hard datasets are displayed in Table \ref{r101}.

\section{Samples of DZEN Questions}
\begin{figure}[htbp]
\centering
\includegraphics[width=\textwidth]{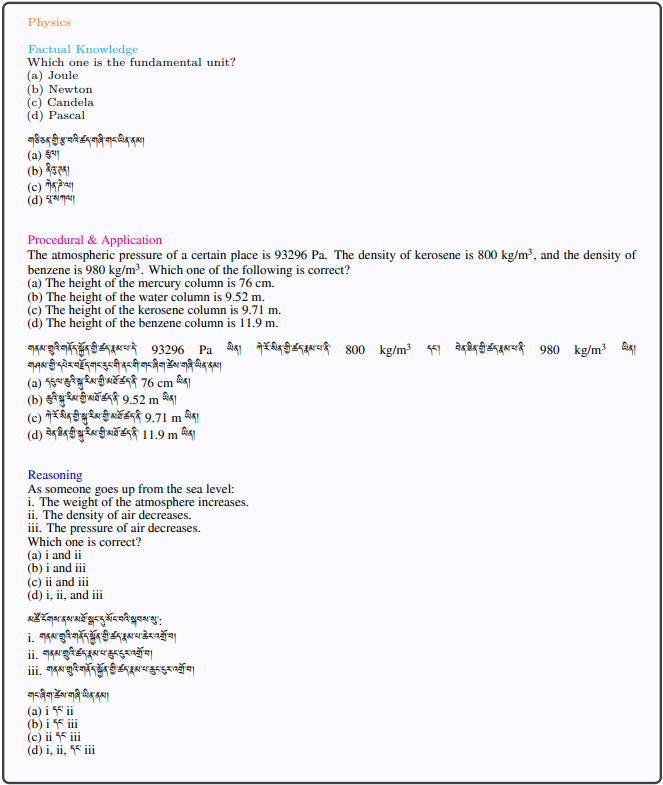}
    \caption{Physics sample questions for the 1oth grade in DZEN by category.}
    \label{f25}
\end{figure}

\begin{figure}[htbp]
\centering
\includegraphics[width=\textwidth]{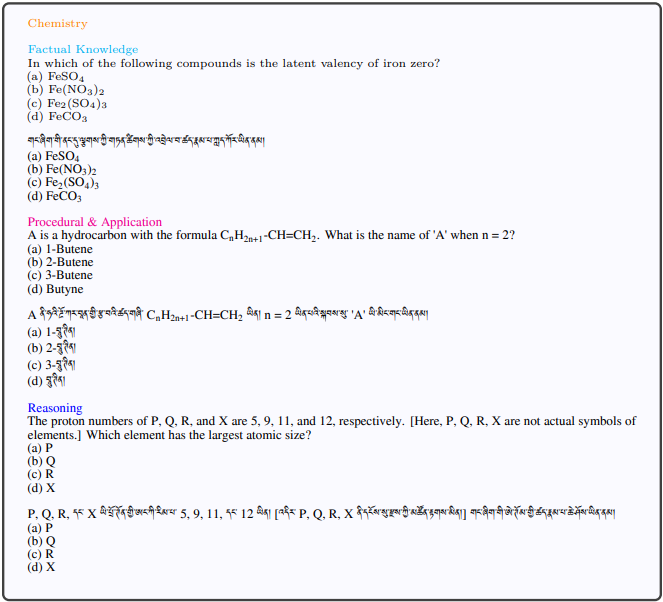}
\caption{Chemistry sample questions for the 1oth grade in DZEN by category.}
    \label{f26}
\end{figure}

\begin{figure}[htbp]
\centering
\includegraphics[width=\textwidth]{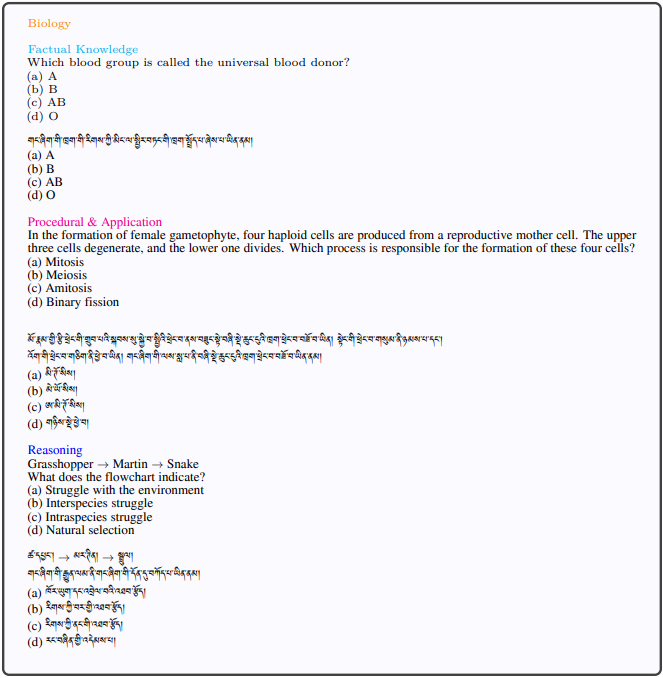}
    \caption{Biology sample questions for the 10th Grade in DZEN by category.}
    \label{f27}
\end{figure}

\begin{figure}[htbp]
\centering
\includegraphics[width=\textwidth]{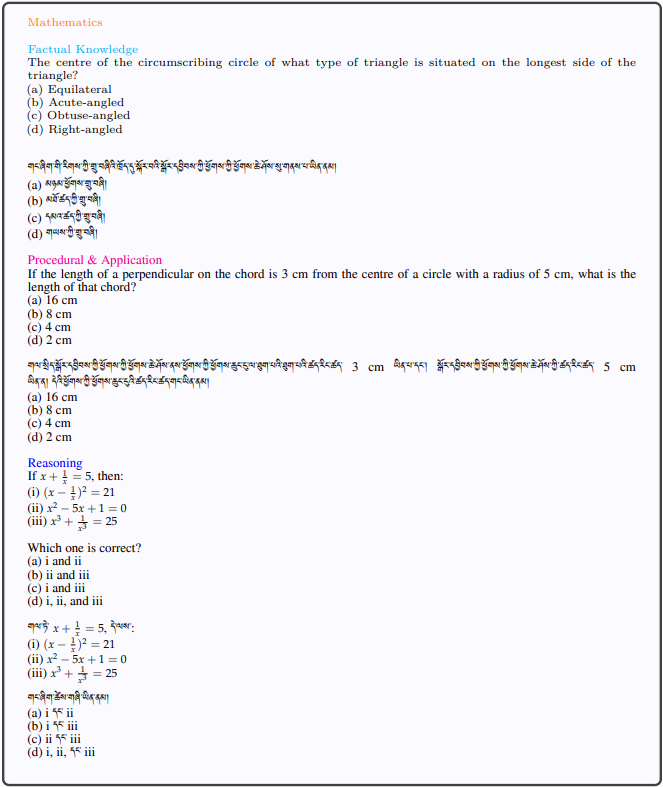}
\caption{Mathematics sample questions for the 1oth grade in DZEN by category.}
    \label{f28}
\end{figure}

Figures \ref{f25}, \ref{f26}, \ref{f27} \& \ref{f28} provide a few examples from the 10th-grade subjects by category, with both English and Dzongkha translations. We left the questions exactly as they are, meaning that we didn't fix the minor grammatical errors in some of the English versions.

\begin{figure}[htbp]
    \centering
    \begin{tcolorbox}[colback=blue!2!white]
    
    You are provided with a multiple-choice exam question written in English, including its answer choices. The question is generally well-constructed but may occasionally contain slight grammatical errors or awkward phrasing. Your task is to revise and polish the question to make it sound more natural and fluent.\\ 
    \break
    The answer choices are given for context, but your response should not include them. Ensure that you address the entire question, not just a portion of it. If there is an equation or formula within the question, leave it unchanged.\\
    Here is the question that needs revision:
    \end{tcolorbox}
    \caption{GPT-4 prompt for creating Grammar-corrected questions.}
    \label{f21}
\end{figure}

\begin{figure}[htbp]
    \centering
    \begin{tcolorbox}[colback=blue!2!white]

    You are given a multiple-choice question, and your task is to determine what type of reasoning is necessary to solve it. Choose from the following options:\\  
    \break
    1. Factual Knowledge: The question relies on recalling basic facts, dates, events, or concepts.\\ 
    2. Procedural and Application: The question requires applying a procedure, formula, or set of steps to reach the answer.\\  
    3. Reasoning: The question demands multistep reasoning or logical deduction to find the solution.\\
    Here is the question for you to evaluate:
    \end{tcolorbox}
    \caption{DZEN dataset categorized questions prompt.}
    \label{f20}
\end{figure}

\begin{figure}[htbp]
    \centering
    \begin{tcolorbox}[colback=blue!2!white]
    You are provided with a multiple-choice question and its options in English or Dzongkha. Your task is to correctly answer the question. Follow these steps:\\ 
    \break
    1. Think and reason step by step, explaining your thought process in either English or Dzongkha.\\  
    2. After reasoning, select the final answer and provide it only as one letter: "a", "b", "c", or "d".\\  
    3. Do not include the text of the option—just the letter of the answer.\\ 
    4. There will always be one correct answer among the given options, so never respond with "none of the above" or give multiple answers.\\  
    Here is the question for you to answer:
    \end{tcolorbox}
    \caption{Proprietary models prompt for DZEN zero-shot benchmark.}
    \label{f22}
\end{figure}

\begin{figure}[htbp]
    \centering
    \begin{tcolorbox}[colback=blue!2!white]
    
    You are provided with a multiple-choice question in either English or Dzongkha. Your task is to solve the question step by step and then give the correct answer.\\
    \break
    Format:\\ 
    \{Question 1\}\\
    \{Reasoning and Explanation\}\\  
    \{Answer 1\}\\  
    \{Question 2\}\\  
    \{Reasoning and Explanation\}\\  
    \{Answer 2\}\\  
    -----------------------------------------------------------------------------------------------------------------------\\  
    \{Question\}\\ 
    Now, follow the same process for the given question.\\
    \end{tcolorbox}
    \caption{Proprietary models prompt for DZEN Few-shot benchmark.}
    \label{f23}
\end{figure}

\begin{figure}[htbp]
    \centering
    \begin{tcolorbox}[colback=blue!2!white]
    
    You are given a situation along with its possible reason or answer in Dzongkha. Your task is to identify the correct reason or answer for the given situation.\\
    \break
    For your convenience, an English translation is also provided. However, you must respond only in Dzongkha.\\ 
    Explain your reasoning step by step in Dzongkha, and finally provide your answer.\\  
    Here is the question:\\
    \end{tcolorbox}
\caption{Zero-shot append experiment prompt.}
    \label{f24}
\end{figure}

\begin{table}[htbp]
\scriptsize
    \centering
    \caption{DZEN dataset question count by subjects and categories.}
    \begin{tabular}{p{4cm}p{4cm}p{2cm}p{2cm}}
        \toprule
        \textbf{Subject} & \textbf{Category} & \textbf{Questions} & \textbf{Instances (\%)} \\
        \midrule
        \multicolumn{4}{c}{\textit{12th Grade Subjects}} \\
        12th Bio I & Factual Knowledge & 288 & 5.58\% \\
                   & Procedural \& Application & 12 & 0.23\% \\
                   & Reasoning & 15 & 0.29\% \\
        12th Bio II & Factual Knowledge & 297 & 5.75\% \\
                   & Procedural \& Application & 8 & 0.16\% \\
                   & Reasoning & 28 & 0.54\% \\
        12th Chem I & Factual Knowledge & 229 & 4.44\% \\
                    & Procedural \& Application & 85 & 1.65\% \\
                    & Reasoning & 58 & 1.12\% \\
        12th Chem II & Factual Knowledge & 185 & 3.59\% \\
                     & Procedural \& Application & 145 & 2.81\% \\
                     & Reasoning & 64 & 1.24\% \\
        12th Phy I & Factual Knowledge & 115 & 2.23\% \\
                   & Procedural \& Application & 161 & 3.12\% \\
                   & Reasoning & 32 & 0.62\% \\
        12th Phy II & Factual Knowledge & 168 & 3.25\% \\
                    & Procedural \& Application & 143 & 2.77\% \\
                    & Reasoning & 27 & 0.52\% \\
        12th Math I & Factual Knowledge & 13 & 0.25\% \\
                    & Procedural \& Application & 368 & 7.13\% \\
                    & Reasoning & 20 & 0.39\% \\
        12th Math II & Factual Knowledge & 24 & 0.46\% \\
                     & Procedural \& Application & 327 & 6.34\% \\
                     & Reasoning & 45 & 0.87\% \\
        \midrule
        \multicolumn{4}{c}{\textit{10th Grade Subjects}} \\
        10th Bio & Factual Knowledge & 308 & 5.97\% \\
                 & Procedural \& Application & 21 & 0.41\% \\
                 & Reasoning & 27 & 0.52\% \\
        10th Phy & Factual Knowledge & 178 & 3.45\% \\
                 & Procedural \& Application & 119 & 2.31\% \\
                 & Reasoning & 27 & 0.52\% \\
        10th Math I & Factual Knowledge & 45 & 0.87\% \\
                   & Procedural \& Application & 267 & 5.17\% \\
                   & Reasoning & 73 & 1.41\% \\
        10th Math II & Factual Knowledge & 24 & 0.46\% \\
                    & Procedural \& Application & 316 & 6.12\% \\
                    & Reasoning & 58 & 1.12\% \\
        10th Chem & Factual Knowledge & 268 & 5.19\% \\
                  & Procedural \& Application & 91 & 1.76\% \\
                  & Reasoning & 35 & 0.68\% \\
        \midrule
        \multicolumn{4}{c}{\textit{8th Grade Subjects}} \\
        8th Math & Factual Knowledge & 30 & 0.58\% \\
                 & Procedural \& Application & 139 & 2.69\% \\
                 & Reasoning & 45 & 0.87\% \\
        8th Sci & Factual Knowledge & 194 & 3.76\% \\
                & Procedural \& Application & 26 & 0.50\% \\
                & Reasoning & 13 & 0.25\% \\
        \midrule
        \textbf{Total} & \textbf{All} & \textbf{5161} & \textbf{100.00\%} \\
        \bottomrule
    \end{tabular}
    \label{tab:DZEN_question_count}
\end{table}

\begin{table}[htbp]
\centering
\scriptsize
\caption{DZEN zero-shot benchmark.}
\label{tab:model_performance_v2}
\begin{tabular}{@{}llrrrrrr@{}}
\toprule
\textbf{Language} & \textbf{Subject} & \textbf{GPT-4} & \textbf{GPT-3.5} & \textbf{Claude 2.1} & \textbf{LLaMA2 (13b)} & \textbf{LLaMA2 (7b)} & \textbf{Mistral 7b} \\
\midrule
\multirow{15}{*}{\textbf{English}} 
& 12th Bio I       & 84.67 & 69.33 & 59.52 & 38.24 & 30.61 & 31.59 \\
& 12th Bio II      & 82.15 & 64.11 & 55.44 & 33.82 & 24.79 & 38.27 \\
& 12th Chem I      & 82.83 & 57.51 & 52.18 & 24.29 & 19.47 & 27.23 \\
& 12th Chem II     & 81.12 & 56.24 & 51.89 & 23.75 & 14.63 & 22.51 \\
& 12th Phy I       & 81.47 & 62.33 & 48.41 & 29.28 & 17.32 & 25.39 \\
& 12th Phy II      & 82.48 & 59.63 & 37.95 & 23.52 & 25.61 & 27.96 \\
& 12th Math I      & 85.79 & 59.27 & 58.12 & 10.37 & 6.35  & 11.38 \\
& 12th Math II     & 77.21 & 53.45 & 56.73 & 17.05 & 6.95  & 15.28 \\
& 10th Bio         & 79.95 & 65.03 & 51.91 & 35.02 & 25.17 & 36.68 \\
& 10th Phy         & 78.88 & 62.63 & 53.19 & 29.07 & 19.53 & 26.92 \\
& 10th Math I      & 86.72 & 61.31 & 48.11 & 13.35 & 12.33 & 12.84 \\
& 10th Math II     & 84.39 & 63.17 & 49.68 & 9.41  & 10.67 & 15.41 \\
& 10th Chem        & 86.78 & 62.59 & 56.97 & 28.83 & 22.24 & 28.58 \\
& 8th Math         & 85.27 & 70.24 & 56.23 & 26.34 & 21.19 & 21.67 \\
& 8th Sci          & 77.43 & 63.27 & 54.24 & 42.23 & 23.78 & 37.08 \\
\midrule
\multirow{15}{*}{\textbf{Dzongkha}}
& 12th Bio I       & 65.23 & 28.57 & 32.14 & \textit{--} & \textit{--} & \textit{--} \\
& 12th Bio II      & 63.87 & 25.34 & 28.93 & \textit{--} & \textit{--} & \textit{--} \\
& 12th Chem I      & 61.42 & 24.16 & 27.53 & \textit{--} & \textit{--} & \textit{--} \\
& 12th Chem II     & 59.85 & 22.63 & 25.82 & \textit{--} & \textit{--} & \textit{--} \\
& 12th Phy I       & 62.58 & 26.43 & 29.75 & \textit{--} & \textit{--} & \textit{--} \\
& 12th Phy II      & 60.17 & 23.85 & 26.27 & \textit{--} & \textit{--} & \textit{--} \\
& 12th Math I      & 58.35 & 21.53 & 24.86 & \textit{--} & \textit{--} & \textit{--} \\
& 12th Math II     & 54.68 & 18.94 & 22.15 & \textit{--} & \textit{--} & \textit{--} \\
& 10th Bio         & 61.73 & 25.67 & 29.27 & \textit{--} & \textit{--} & \textit{--} \\
& 10th Phy         & 59.28 & 23.17 & 26.85 & \textit{--} & \textit{--} & \textit{--} \\
& 10th Math I      & 57.46 & 20.83 & 23.57 & \textit{--} & \textit{--} & \textit{--} \\
& 10th Math II     & 55.94 & 19.47 & 21.96 & \textit{--} & \textit{--} & \textit{--} \\
& 10th Chem        & 60.35 & 24.37 & 27.64 & \textit{--} & \textit{--} & \textit{--} \\
& 8th Math         & 63.57 & 27.28 & 30.17 & \textit{--} & \textit{--} & \textit{--} \\
& 8th Sci          & 58.47 & 22.75 & 25.48 & \textit{--} & \textit{--} & \textit{--} \\
\bottomrule
\end{tabular}
\label{tab5}
\end{table}

\begin{table}[htbp]
\centering
\scriptsize
\caption{DZEN 10th grade few-shot benchmark. Note that w/ denotes with and w/o denotes without CoT.}
\label{tab:combined_cot}
\begin{tabular}{@{}p{1cm}p{1.5cm}p{1cm}p{1cm}p{1cm}p{1.5cm}p{1.5cm}p{1.5cm}p{1cm}@{}}
\toprule
\textbf{Language} & \textbf{Dataset} & \textbf{CoT} & \multicolumn{2}{c}{\textbf{GPT 3.5}} & \multicolumn{2}{c}{\textbf{LLaMA2 (7b)}} & \multicolumn{2}{c}{\textbf{LLaMA2 (13b)}} \\
\cmidrule(lr){4-5} \cmidrule(lr){6-7} \cmidrule(lr){8-9}
 & & & \textbf{5-shot} & \textbf{3-shot} & \textbf{5-shot} & \textbf{3-shot} & \textbf{5-shot} & \textbf{3-shot} \\
\midrule
\multirow{5}{*}{English} 
& \multirow{2}{*}{10th Bio} & w/o & 58.39 & 59.27 & 48.85 & 43.63 & 43.63 & 39.28 \\
& & w/ & 67.95 & 69.63 & 49.72 & 49.72 & 48.85 & 43.63 \\
\cmidrule(lr){2-9}
& \multirow{2}{*}{10th Phy} & w/o & 56.67 & 61.02 & 37.52 & 39.28 & 29.72 & 27.98 \\
& & w/ & 66.24 & 70.58 & 40.15 & 38.41 & 38.41 & 43.63 \\
\cmidrule(lr){2-9}
& \multirow{2}{*}{10th Math I} & w/o & 41.89 & 43.63 & 30.58 & 25.37 & 34.06 & 28.85 \\
& & w/ & 66.24 & 69.63 & 28.85 & 26.24 & 37.52 & 33.19 \\
\cmidrule(lr){2-9}
& \multirow{2}{*}{10th Math II} & w/o & 46.24 & 37.52 & 21.89 & 21.89 & 40.15 & 42.76 \\
& & w/ & 62.76 & 58.39 & 29.72 & 27.98 & 25.37 & 30.58 \\
\cmidrule(lr){2-9}
& \multirow{2}{*}{10th Chem} & w/o & 58.39 & 58.39 & 36.67 & 37.52 & 27.11 & 25.37 \\
& & w/ & 68.85 & 63.63 & 44.50 & 42.76 & 43.63 & 42.76 \\
\midrule
\multirow{5}{*}{Dzongkha} 
& \multirow{2}{*}{10th Bio} & w/o & 28.14 & 27.42 & \textit{--} & \textit{--} & \textit{--} & \textit{--} \\
& & w/ & 31.12 & 28.89 & \textit{--} & \textit{--} & \textit{--} & \textit{--} \\
\cmidrule(lr){2-9}
& \multirow{2}{*}{10th Phy} & w/o & 35.52 & 30.32 & \textit{--} & \textit{--} & \textit{--} & \textit{--} \\
& & w/ & 39.21 & 33.28 & \textit{--} & \textit{--} & \textit{--} & \textit{--} \\
\cmidrule(lr){2-9}
& \multirow{2}{*}{10th Math I} & w/o & 22.94 & 27.42 & \textit{--} & \textit{--} & \textit{--} & \textit{--} \\
& & w/ & 29.65 & 35.52 & \textit{--} & \textit{--} & \textit{--} & \textit{--} \\
\cmidrule(lr){2-9}
& \multirow{2}{*}{10th Math II} & w/o & 28.14 & 28.85 & \textit{--} & \textit{--} & \textit{--} & \textit{--} \\
& & w/ & 34.78 & 38.47 & \textit{--} & \textit{--} & \textit{--} & \textit{--} \\
\cmidrule(lr){2-9}
& \multirow{2}{*}{10th Chem} & w/o & 35.52 & 35.52 & \textit{--} & \textit{--} & \textit{--} & \textit{--} \\
& & w/ & 34.78 & 40.68 & \textit{--} & \textit{--} & \textit{--} & \textit{--} \\
\bottomrule
\end{tabular}
\label{tab6}
\end{table}

\begin{table}[htbp]
\centering
\scriptsize
\caption{Big-Bench Hard zero-shot benchmark.}
\label{tab:dzongkha_benchmark}
\begin{tabular}{@{}p{3cm}p{2cm}p{2cm}p{3cm}p{3cm}@{}}
\toprule
\textbf{Dataset} & \textbf{English Only} & \textbf{Dzongkha Only} & \textbf{Translation Append (Premium)} & \textbf{Translation Append (GPT)} \\
\midrule
Causal Judgement & 57.2 & 43.2 & 53.0 & 51.4 \\
Date Understanding & 56.0 & 20.2 & 36.8 & 28.1 \\
Disambiguation QA & 45.2 & 32.0 & 47.4 & 45.9 \\
Formal Fallacies & 55.2 & 42.9 & 42.8 & 41.4 \\
Logical Deductions Five & 35.2 & 17.3 & 15.1 & 16.6 \\
Logical Deductions Seven & 30.8 & 13.1 & 18.2 & 13.4 \\
Logical Deductions Three & 51.6 & 28.2 & 30.7 & 31.2 \\
Multistep Arithmetic & 68.5 & 39.2 & 34.8 & -- \\
Navigate & 60.4 & 44.5 & 49.7 & 50.8 \\
Object Counting & 83.6 & 27.8 & 34.9 & 37.4 \\
Penguin In A Table & 61.0 & 20.8 & 29.1 & 29.1 \\
Reasoning About Colored Objects & 42.4 & 19.2 & 25.9 & 23.0 \\
Temporal Sequences & 37.6 & 18.9 & 18.2 & 17.0 \\
Web of Lies & 52.8 & 17.0 & 39.0 & 38.4 \\
\bottomrule
\end{tabular}
\label{r101}
\end{table}

\end{document}